\documentclass{article}

    \PassOptionsToPackage{numbers, sort, compress}{natbib}

\usepackage[preprint]{neurips_2022}

\usepackage[utf8]{inputenc} 
\usepackage[T1]{fontenc}    
\usepackage{url}            
\usepackage{booktabs}       
\usepackage{amsfonts}       
\usepackage{nicefrac}       
\usepackage{microtype}      
\usepackage[dvipsnames]{xcolor}         
\usepackage[colorlinks=true, citecolor=RoyalBlue, linkcolor=magenta]{hyperref}
\usepackage{graphicx}
\usepackage{amsmath,amssymb} %
\usepackage{subfig}
\usepackage{enumitem}
\usepackage[font=small]{caption}

\usepackage{algorithm}
\usepackage{algorithmic}
\usepackage{bbding}
\usepackage{colortbl}
\usepackage{listings}
\usepackage[dvipsnames]{xcolor}
\usepackage{hyperref}
\usepackage{xspace}
\usepackage{multirow}
\usepackage{amssymb}
\usepackage{tablefootnote}
\usepackage{etoolbox}
\usepackage{stmaryrd} 
\usepackage{enumitem}

\makeatletter
\DeclareRobustCommand{\shortto}{%
  \mathrel{\mathpalette\short@to\relax}%
}

\newcommand{\short@to}[2]{%
  \mkern2mu
  \clipbox{{.5\width} 0 0 0}{$\m@th#1\vphantom{+}{\shortrightarrow}$}%
  }
\makeatother

\usepackage{pifont}
\newcommand{\cmark}{\ding{51}}
\newcommand{\xmark}{\ding{55}}

\usepackage{cleveref}
\crefformat{section}{\S#2#1#3} 
\crefformat{subsection}{\S#2#1#3}
\crefformat{subsubsection}{\S#2#1#3}

\usepackage{xspace}

\newcommand\model{\textsc{MimDet}}

\renewcommand{\paragraph}[1]{\vspace{1.25mm}\noindent\textbf{#1}}

\makeatletter
\DeclareRobustCommand\onedot{\futurelet\@let@token\@onedot}
\def\@onedot{\ifx\@let@token.\else.\null\fi\xspace}

\def\eg{\emph{e.g}\onedot} 
\def\ie{\emph{i.e}\onedot} 
 
 \def\vs{\emph{vs}\onedot}

\makeatother

\makeatletter
\AfterEndEnvironment{algorithm}{\let\@algcomment\relax}
\AtEndEnvironment{algorithm}{\kern2pt\hrule\relax\vskip3pt\@algcomment}
\let\@algcomment\relax
\newcommand\algcomment[1]{\def\@algcomment{\footnotesize#1}}
\renewcommand\fs@ruled{\def\@fs@cfont{\bfseries}\let\@fs@capt\floatc@ruled
  \def\@fs@pre{\hrule height.8pt depth0pt \kern2pt}%
  \def\@fs@post{}%
  \def\@fs@mid{\kern2pt\hrule\kern2pt}%
  \let\@fs@iftopcapt\iftrue}
\makeatother

\makeatletter
\renewcommand*{\ALG@name}{Architecture}
\makeatother

\newcolumntype{x}[1]{>{\centering\arraybackslash}p{#1pt}}
\newcolumntype{y}[1]{>{\raggedright\arraybackslash}p{#1pt}}
\newcolumntype{z}[1]{>{\raggedleft\arraybackslash}p{#1pt}}
\newlength\savewidth\newcommand\shline{\noalign{\global\savewidth\arrayrulewidth
  \global\arrayrulewidth 1pt}\hline\noalign{\global\arrayrulewidth\savewidth}}

\definecolor{gain}{HTML}{34a853}  %

\definecolor{lost}{HTML}{ea4335}  %

\makeatletter\renewcommand\paragraph{\@startsection{paragraph}{4}{\z@}
  {.5em \@plus1ex \@minus.2ex}{-.5em}{\normalfont\normalsize\bfseries}}\makeatother

\definecolor{baselinecolor}{gray}{.9}
\newcommand{\baseline}[1]{\cellcolor{baselinecolor}{#1}}
\newcommand{\boxAP}{AP$^\text{box}$\xspace}
\newcommand{\maskAP}{AP$^\text{mask}$\xspace}

\title{Unleashing Vanilla Vision Transformer with \\ Masked Image Modeling for Object Detection}

\author{%
  \vspace{0.4em}
  Yuxin Fang$^{1}$\thanks{Equal contribution.
  $^\dag$Corresponding author (\texttt{xgwang@hust.edu.cn}).
  This work was done when Shusheng Yang was interning at ARC Lab, Tencent PCG.} \ \
  Shusheng Yang$^{1*}$ \
  Shijie Wang$^{1*}$ \
  Yixiao Ge$^{2}$ \
  Ying Shan$^{2}$ \
  Xinggang Wang$^{1\dag}$
  \\
  \vspace{.15em}
  $^{1}$ School of EIC, Huazhong University of Science \& Technology
  \\
  \vspace{.15em}
  $^{2}$ ARC Lab, Tencent PCG
}

\begin{document}

\maketitle

\setcounter{footnote}{0}

\definecolor{convcolor}{HTML}{412F8A}
\definecolor{resnetcolor}{HTML}{8DA0CB}
\definecolor{vitcolor}{HTML}{fc8e62}

\newcommand{\convcolor}[1]{\textcolor{convcolor}{#1}}
\newcommand{\vitcolor}[1]{\textcolor{vitcolor}{#1}}
\newcommand{\cnn}{ConvNeXt}

\newcommand{\hvits}{\vitcolor{$\bullet$\,}}
\newcommand{\hvite}{\vitcolor{$\mathbf{\circ}$\,}}
\newcommand{\vvits}{\convcolor{$\bullet$\,}}
\newcommand{\vvite}{\convcolor{$\mathbf{\circ}$\,}}
\newcommand{\gr}{\rowcolor[gray]{.95}}
    
\begin{abstract}
We present an approach to efficiently and effectively adapt a masked image modeling (MIM) pre-trained vanilla Vision Transformer (ViT) for object detection, which is based on our two novel observations: (i) A MIM pre-trained vanilla ViT encoder can work surprisingly well in the challenging object-level recognition scenario even with \textit{randomly sampled partial} observations, \eg{}, only 25\% $\sim$ 50\% of the input embeddings. (ii) In order to construct multi-scale representations for object detection from single-scale ViT, a \textit{randomly initialized compact convolutional} stem supplants the pre-trained large kernel patchify stem, and its intermediate features can naturally serve as the higher resolution inputs of a feature pyramid network without further upsampling or other manipulations. While the pre-trained ViT is only regarded as the 3$^{rd}$-stage of our detector's backbone instead of the whole feature extractor. This results in a ConvNet-ViT \textit{hybrid} feature extractor. The proposed detector, named \model{}, enables a MIM pre-trained vanilla ViT to outperform hierarchical Swin Transformer by 2.5 \boxAP and 2.6 \maskAP on COCO, and achieves better results compared with the previous best adapted vanilla ViT detector using a more modest fine-tuning recipe while converging 2.8$\times$ faster. Code and pre-trained models are available at \url{https://github.com/hustvl/MIMDet}.
\end{abstract}

\section{Introduction}

Transformer~\cite{Transformer} is born to transfer.
Entering the 2020s, Vision Transformer (ViT)~\cite{ViT} is rapidly transferring the viewpoint of machine vision in both architectural design~\cite{Swin, PVT, MViT, CoaT, pit} as well as representation learning~\cite{beit, mae, simmim, ibot, cim}.
From the perspective of architectural design, Swin Transformer~\cite{Swin}, as a representative, seamlessly incorporates the local window attention into a hierarchical macro architecture.
The window attention enables Swin Transformer to efficiently process high-resolution inputs with feasible costs, and the inherent pyramidal feature hierarchy can better handle the large variations of visual entities' scales and sizes, which is crucial for object-level understanding.
Meanwhile, from the perspective of general visual representation learning, masked image modeling (MIM)~\cite{beit, mae} pre-trained vanilla ViT\footnote{For disambiguation, in this paper, ``vanilla ViT'' refers to the Vision Transformer proposed by~\citet{ViT} with the isotropic, single-scale architecture unless specified.} demonstrates promising scalability and is superior to the supervised isotropic or hierarchical ViT counterparts when transferred to image as well as video classification tasks~\cite{ibot, peco, maskfeat, data2vec} with low-resolution inputs. 
 
The compelling transfer learning performance of MIM pre-trained ViTs in image- and video-level recognition tasks motivate us to ponder: Is it possible for the more challenging \textit{object} or \textit{instance}-level recognition tasks, \eg{}, object detection and instance segmentation, to also benefit from the powerful MIM pre-trained representations?
Unfortunately, it is quite slow for vanilla ViT to directly process high-resolution images required by object-level recognition, for the complexity of global attention scales quadratically with the spatial dimension.
One way is to re-introduce window attention or its variants during the fine-tuning stage, which can help reduce the attention's costs~\cite{benchmarking}.
However, using window attention causes a \textit{discrepancy} between MIM pre-training and fine-tuning, as the window partition implicitly treats inputs as \textit{2D continuous regular grids}, while the vanilla ViT is pre-trained to process \textit{1D partial sequences}.

\begin{figure}[t!]
\vspace{-1.5em}
    \newcommand{\sz}{0.5}
    \makebox[\textwidth][c]{
    \begin{minipage}{1.08\columnwidth}  %
    \centering
    \includegraphics[height=\sz\linewidth]{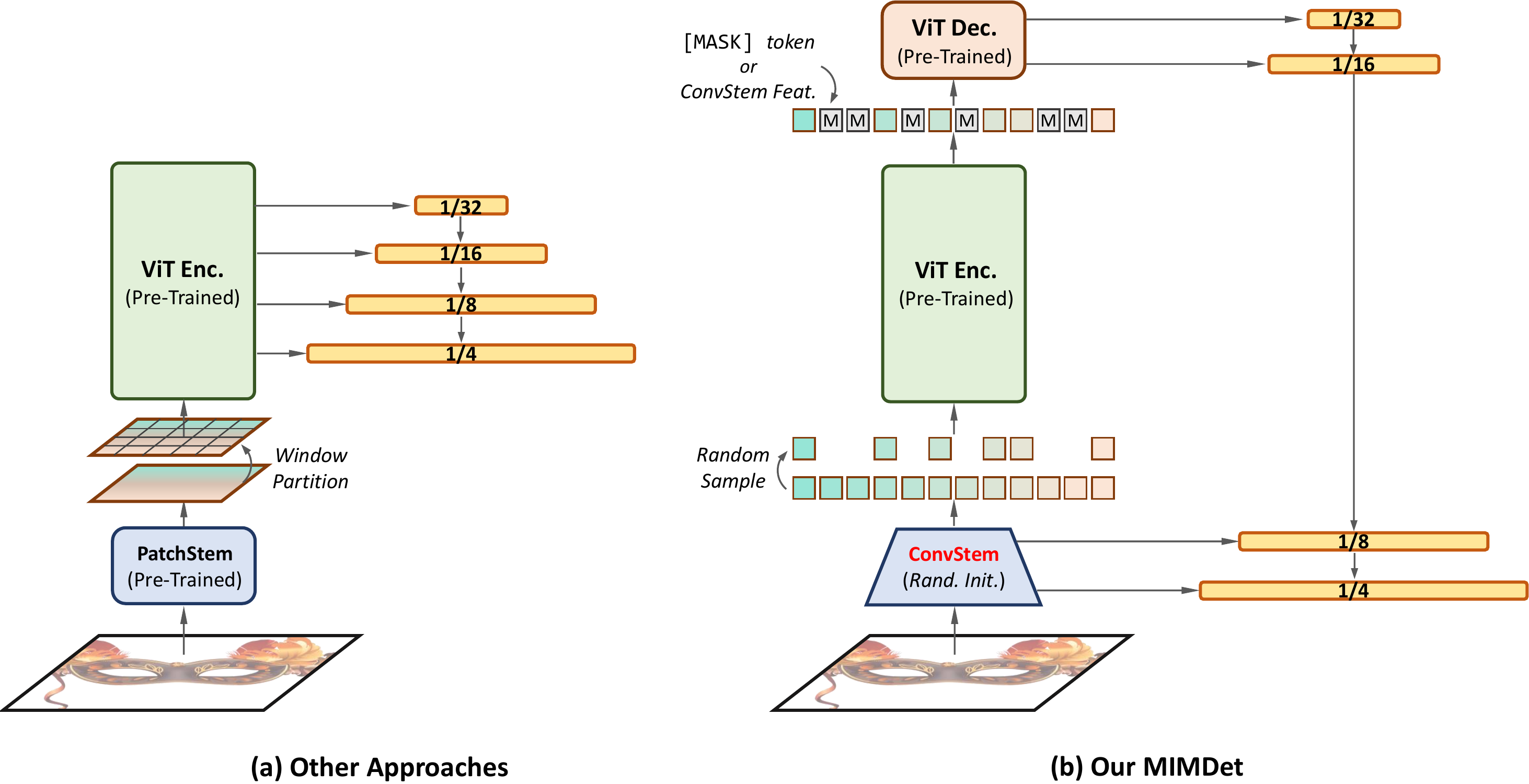}
    \end{minipage}
    }
  \caption{Overview of our \model{} and comparisons with previous representative approaches adapting vanilla ViT for object detection (\eg{}, ~\citet{benchmarking}). In \model{}, a randomly initialized compact convolutional stem (ConvStem) replaces the pre-trained large kernel patchify stem (PatchStem), and the ViT encoder only receives and processes sampled partial input embeddings. The intermediate features of ConvStem can directly serve as the higher resolution inputs for a standard FPN~\cite{FPN}.}
\vspace{-1.5em}
  \label{fig: mimdet}
\end{figure}


Inspired by the MIM pre-training~\cite{beit, mae}, this work pursues a radical solution to transfer a vanilla ViT for object-level recognition (Figure~\ref{fig: mimdet}b):
We feed the MIM pre-trained ViT encoder with \textit{only a partial input}, \eg{}, only 25\% $\sim$ 50\% of the input sequence of embeddings with \textit{random sampling}, during the object detection fine-tuning stage. 
The output sequence fragments are then complemented with learnable tokens (\eg, $\mathtt{[MASK]}$ tokens) and processed by a lightweight decoder to recover the full feature map.
This seemingly bold approach turns out to be surprisingly good in terms of accuracy-resource trade-off.
Our motivation is that: 
(i) For a long time, the 2D \textit{continuous} regular grid is considered as the \textit{de facto} input from of Convolutional Neural Networks (ConvNets)~\cite{lecun1989backpropagation} as well as hierarchical ViTs in visual understanding.
Unlike them, vanilla ViT treats the input as a sequence of individual tokens / embeddings. 
Therefore it is feasible for ViT to process \textit{nonconsecutive} input subsets, which serves as the foundation of our approach.
(ii) Visual signal has heavy spatial redundancy by nature, which encourages the recent MIM pre-training approaches to adopt a very high masking ratio (\eg{}, 40\% $\sim$ 75\%)~\cite{beit, mae}. 
The successful practice of generative MIM pre-training implies that it is also possible to perform holistic visual understanding or recognition given only an input subset, as we believe visual generation and recognition are two sides of the same coin in essence.
Since our approach also encourages ViT to implicitly conduct a kind of MIM during fine-tuning, it introduces a smaller gap between upstream MIM pre-training and downstream detector fine-tuning compared with the window attention and its variants.

Another obstacle is the lack of pyramidal feature hierarchy in vanilla ViT, as well-established visual recognition task layers~\cite{FPN, RetinaNet, MaskRCNN, EfficientDet} usually require multi-scale inputs.
In this work, we don't aim to re-design specific task layers tailored for single-scale ViT, instead, we make minimal adaptations to re-introduce multi-scale representations from vanilla ViT for better leveraging the precious legacy of visual understanding study.
Different from previous approaches, which treats the vanilla ViT as the \textit{whole} feature extractor / backbone and artificially manipulates intermediate features to compromise with FPN~\cite{FPN}, we consider ViT only as a \textit{part} of a hierarchical backbone, \ie{}, the 3$^{rd}$-stage, and a \textit{randomly initialized neat convolutional} stem (ConvStem) supplants the pre-trained large kernel patchify stem as the 1$^{th}$- and 2$^{nd}$-stage.
The newly introduced ConvStem is very compact (only 4.1M, less than 5\% of the ViT-Base encoder's size), and its intermediate features can directly serve as the higher resolution inputs of a feature pyramid network.
The resulting feature extractor is essentially a ConvNet-ViT \textit{hybrid} architecture with a neat ConvStem as the earlier stage and a strong ViT as the later stage, which is also consistent with the current trends of the state-of-the-art visual encoder design methodology that tries to combine the strengths of two kinds of layers~\cite{HaloNet, coatnet}.

The odyssey of object detection research is like ``a goose that laid the golden eggs'' that can always inspire and generalize well to the study of other visual understanding tasks.
Therefore we choose the ubiquitous \& canonical Mask R-CNN~\cite{MaskRCNN} as a touchstone for our design.
The detector, named \model{} (\textbf{M}asked \textbf{I}mage \textbf{M}odeling for \textbf{Det}ection, Figure~\ref{fig: mimdet}b), enables a MIM pre-trained vanilla ViT-Base model (128M) to obtain 51.7 \boxAP and 46.1 \maskAP on the COCO dataset~\cite{COCO}, outperforming the hierarchical Swin Transformer counterpart pre-trained on ImageNet-1K~\cite{ImageNet-1k} with supervision by 2.5 \boxAP and 2.6 \maskAP.
Moreover, our approach can achieve even better results compared with the previous best adapted vanilla ViT detector~\cite{benchmarking} based on the same MIM pre-trained representation~\cite{mae} using a more modest fine-tuning recipe while being 2.8$\times$ faster in terms of converge speed.
We also observe a promising scaling trend of our approach.

Our study indicates that \textit{designing} and \textit{pre-training} specific feature extractors for visual recognition may no longer be \textit{indispensable} given the strong representation inside vanilla ViT, as long as we find the right way to unleash it.
We believe a trend in future machine vision research is to better leverage the pre-trained representation via delicately adapting it using simple task layers.
These observations are also aligned with those witnessed in natural language processing (NLP)~\cite{GPT, BERT, t5, CLIP}, and we hope this work can encourage the vision community to explore a similar trajectory.

\section{Method}

The goal is to tame and unleash a MIM pre-trained vanilla ViT to achieve favorable performance in object-level recognition with feasible costs. 
To this end, we propose to (i) feed ViT with only the sampled partial inputs instead of the full input set, described in \cref{sec: yolops}, and (ii) use a randomly initialized compact convolutional stem to replace the pre-trained patchify stem for a better pyramidal feature hierarchy, detailed in \cref{sec: convstem}.
Figure~\ref{fig: mimdet}b illustrates our approach.

\subsection{You Only Look at One \textit{Partial} Sequence}
\label{sec: yolops}

Object-level recognition tasks usually benefit from higher resolution inputs, which are in general over one order of magnitude higher than the input size of image classification.
While the vanilla ViT computes global attention for spatial features aggregation, and the computational and memory costs of global attention scale quadratically with the input resolutions.
Hence the fine-tuning process will be largely slowed down if a vanilla ViT is directly fed with the full input set.

We notice that vanilla ViT treats the input as a sequence of individual elements, therefore it is possible for a vanilla ViT to receive and process only a \textit{partial}, \textit{discontinuous} input sub-sequence given positional information (via position embeddings), which is intrinsically different from ConvNets as well as hierarchical ViT counterparts that manipulating on 2D \textit{continuous} gird inputs.
This property of vanilla ViT encourages us to act bold: we \textit{randomly sample} a subset of patch embeddings serving as the input set of a MAE~\cite{mae} pre-trained vanilla ViT encoder, \ie{}, the encoder only looks at one \textit{partial} input sequence.
Surprisingly, we find that with only 25\% of the input sequence for the ViT-Base encoder, our detector can already achieve a very competitive accuracy that outperforms an augmented Swin Transformer under the same fine-tuning procedure.
Furthermore, with 50\% of the input, \model{} can yield 51.7 \boxAP / 46.1 \maskAP, outperforming Swin by 2.5 \boxAP / 2.6 \maskAP.

To our knowledge, there is little literature to demonstrate that the challenging object-level recognition can be successfully done with only randomly sampled partial inputs.
Our motivation is: 
\begin{enumerate}[label=(\roman*), topsep=0pt, itemsep=0ex, partopsep=1ex, parsep=1ex]
\item The visual signal is highly redundant and loosely spans the spatial dimension, \eg{}, a vanilla ViT is able to recover the missing contextual information based only on a small set of visible content during MIM pre-training~\cite{beit, mae, simmim}, which implies it \textit{understands} the \textit{global} context before generation.
Therefore it is possible to perform global visual understanding in complex scenes based on strong pre-trained representations given only partial observations.
\item Our approach introduces a smaller gap between MIM pre-training and fine-tuning, \ie{} we fine-tune the ViT in a similar vein as pre-training.
As during pre-training, ViT learns a pretext task that requires global contextual reasoning with only a visible subset as inputs. 
Our fine-tuning process mimics the MIM pre-training that conducts global object-level understanding with only partial observations based on high-capacity representations.
\end{enumerate}
This solution enables us to achieve a win-win scenario: it optimizes the accuracy-resource trade-off while introducing a smaller pre-training \& fine-tuning gap as well as leveraging the pre-trained representations more judiciously.
The output sequence fragments are then complemented with learnable tokens (special $\mathtt{[MASK]}$ tokens or feature embeddings, please refer to Table~\ref{tab: mask or conv} for details), and processed by a lightweight MAE pre-trained ViT decoder (\eg, only 4$\times$ Transformer layers with reduced embedding dimension) to recover the full image feature.

Another way to adapt vanilla ViT for high-resolution input is to re-introduce window attention or its variants during the fine-tuning stage~\cite{benchmarking}, which can reduce the attention's compute.
More importantly, this kind of adaptation still treats the input as 2D continuous regular grids as ConvNets and hierarchical ViTs therefore also introducing 2D inductive biases, which is in principle beneficial to object- / region-level understanding.
We also agree that the fine-tuning process would benefit from task-specific prior knowledge, especially given relatively less data available than pre-training.
However, using window attention causes a discrepancy between pre-training and fine-tuning to some extent, as the vanilla ViT is pre-trained to process 1D partial sequences and the window partition along with its inductive biases never appears in the pre-training stage.
In the next section, we present a smarter way to inject 2D inductive biases into our detector.


\subsection{You Only Pre-train the \textit{Third} Stage}
\label{sec: convstem}

\paragraph{Introducing ConvStem for Hierarchical Features Construction.}

Well-established visual understanding task layers usually receive \textit{multi-scale} features as inputs to deal with the large scale and size variations of objects, while the MIM pre-trained \textit{single-scale} vanilla ViT demonstrates compelling scalability and transfer learning performances in the image- and video-level recognition tasks compared with its hierarchical counterparts~\cite{ViT, beit, mae, maskfeat}.
Therefore it is quite promising for an object detector to enjoy the best of two worlds. 
However, the multi-scale features do not naturally exist in a vanilla ViT.
If we can find a sensible way to re-introduce the pyramidal feature hierarchy for it, then re-design specific task layers tailored for the single-scale ViT is no longer needed, and the heritage of visual recognition study can be largely inherited.

Trace to its source, the lack of the feature hierarchy in vanilla ViT is rooted in its early visual signal processing, which is too aggressive for a detector:
ViT ``patchifies'' the input image into 16$\times$16 non-overlapping patches and then embeds them to form the Transformer encoder’s input.
The downsampling rate is so high that a visual entity of interest in the input image smaller than 16$\times$16 pixels will no longer exist in the spatial dimension of Transformer encoder’s input embeddings.
Artificially dividing vanilla ViT into multiple stages and upsampling the downsampled intermediate feature~\cite{xcit, benchmarking} is kind of a compromise to construct a feature pyramid for the detector, since there is no explicit evidence that those disappeared visual entities as well as the low-level details in the spatial dimension will re-appear in their original location faithfully (please also refer to Figure~\ref{fig: feat vis}), especially when typical region-based detectors extract and process object's features based on spatial feature alignment as well as translation \& scale equivariance~\cite{Faster-R-CNN, MaskRCNN}.

To mitigate the aforementioned issues, our solution is to \textit{throw away} the pre-trained large-stride patchify stem, and use a \textit{randomly initialized neat convolutional} stem (ConvStem) as a replacement.
We adopt a minimalist ConvStem design by simply stacking 3$\times$3 convolutions with a stride of 2 and doubled feature dimensions.
Each convolutional layer is followed by a layer normalization~\cite{LN} and a GELU activation~\cite{GELU}.
The detailed architectural configurations are given in the Appendix.
Our ConvStem progressively reduces the spatial dimension as well as enriches the channel dimension.
The output embedding, which has the same shape as the original patchified embedding, serves as the input (before the random sampling process described in \cref{sec: yolops}) for the ViT encoder.

The newly introduced ConvStem is very compact, in terms of the model size, our ConvStem only has 4.1M parameters, which is less than 5\% of the ViT-Base encoder's size.
Small as it is, the pyramidal feature hierarchy naturally exists in our ConvStem's intermediate layers.
We select the features with strides of \{4, 8\} pixels with respect to the input image as the input features of a standard FPN's first two stages (\ie{}, the input of $P_2$ and $P_3$)~\cite{FPN}, while the output of pre-trained ViT decoder (with a stride of 16, detailed in \cref{sec: yolops}) serves as the input of FPN's $P_4$, and the input for $P_5$ is simply obtained via a parameter-less mean pooling upon the output of ViT decoder.
Now, we successfully obtain a feature pyramid network for object detection.

\paragraph{A ConvNet-ViT Hybrid Architecture.}
Previous attempts regard the pre-trained ViT as the \textit{whole} feature extractor~\cite{xcit, benchmarking}, while we treat the ViT as only a \textit{part} of it, \ie{}, the $3^{rd}$-stage.
In essence, our feature extractor turns out to be a ConvNet-ViT \textit{hybrid} architecture with a shallow \& neat ConvNet / ConvStem as the earlier stage and a deep \& strong ViT as the later stage.
This is also consistent with current trends of the state-of-the-art visual encoder design methodology that tries to combine the strengths of two kinds of architectures~\cite{HaloNet, coatnet}, \ie, the ConvNet / ConvStem is more suitable for early visual signal processing, and introduces 2D inductive biases for the ViT encoder \& detector, while the single-scale vanilla ViT is more scalable and tends to have a larger model capacity.

Notice that our ConvStem is used only during fine-tuning and does not need to be pre-trained, which is different from \citet{xiao2021early}.
In fact, the ConvStem cannot be pre-trained via MIM and also does not support MIM pre-training of any visual encoder, since dense-slid convolutions with kernel size larger than stride propagate information across tokens, causing information leakage and impeding the MIM.
This work shows that the pre-trained early stage is not a \textit{sine qua non} for a detector to achieve favorable performances during fine-tuning, \ie{}, you only need to pre-train the $3^{rd}$-stage.

\section{Experiment}
\label{sec: experiment}
\renewcommand{\ttdefault}{ptm}  %
\newcommand{\tablestyle}[2]{\ttfamily\setlength{\tabcolsep}{#1}\renewcommand{\arraystretch}{#2}\centering\footnotesize}

\paragraph{General Setting.}
We conduct our experiments on the COCO dataset~\cite{COCO} using the $\mathtt{Detectron2}$ library~\cite{detectron2}.
Models are trained on the $\mathtt{train2017}$ split and evaluated on the $\mathtt{val2017}$ split.
For \model{}, we perform experiments mainly on ViT-Base / \model{}-Base model using 32$\times$ 32G V100 GPUs with a total batch size of 64 optimized by AdamW~\cite{AdamW} with a learning rate of 8e-5.
We initialize the vanilla ViT part via MAE~\cite{mae} pre-trained weight on ImageNet-1K~\cite{ImageNet-1k}.
An augmented Mask R-CNN~\cite{MaskRCNN} with FPN~\cite{FPN} is chosen as the detection task layer following~\citet{benchmarking}.
Since the vanilla ViT encoder is already pre-trained while the task layer is trained from scratch, the learning rate of the ViT encoder part is divided by a factor of 2 and the learning rate for the task layer is multiplied by 2.
We use a modest fine-tuning recipe following Swin Transformer~\cite{Swin}, which is a 36-epoch schedule using multi-scale training (scale the shorter side in [480, 800] while the longer side is smaller than 1333) and random crop augmentation.
``pt'' \& ``ft'' means pre-training \& fine-tuning.
The detailed settings and configurations are in the Appendix.

For the evaluation metrics, we report \boxAP for object detection and \maskAP for instance segmentation, with a particular focus on the fine-tuning epochs / wall-clock time, as we care about \textit{how efficiently} a set of \textit{general upstream} visual representations can be adapted to a \textit{specific downstream} task.


\subsection{Study and Analysis}
\label{sec: study and analysis}
We study and analyze the main properties of \model{}-Base via ablating the \colorbox{baselinecolor}{default configuration}.

\begin{table}[ht]
\vspace{-.5em}
    \tablestyle{1.5pt}{1.25}
    \begin{tabular}{@{}z{72}|x{48}x{48}x{48}x{48}x{48}|y{24}y{24}}
        training \ \ \ \ \ & \multicolumn{5}{c|}{inference sampling ratio ($\downarrow$)} & \multicolumn{2}{c}{training} \\
        sampling ratio ($\downarrow$) \ \ & 12.5\% & 25\% & 50\% & 75\% & \baseline{100\%} & \ time & mem$^\dagger$ \\
        \shline
        12.5\% \ \ \ \ \ & 43.0 / 38.7 & 46.1 / 41.7 & 47.5 / 42.9 & 47.9 / 43.2 & 47.6 / 43.0 & \ \scriptsize{16.0 h} & \scriptsize{14.2 G} \\
        25\% \ \ \ \ \   & & 46.8 / 42.2 & 49.4 / 44.2 & 50.0 / 44.7 & 49.9 / 44.7 & \ \scriptsize{16.5 h} & \scriptsize{15.9 G} \\
        \baseline{50\%} \ \ \ \ \  & & & 49.5 / 44.3 & 51.0 / 45.2 & \baseline{51.5 / 46.0} & \ \scriptsize{20.3 h} & \scriptsize{19.9 G} \\
        75\% \ \ \ \ \   & & & & 51.0 / 45.6 & 51.8 / 46.3 & \ \scriptsize{27.2 h} & \scriptsize{28.5 G} \\
        100\% \ \ \ \ \  & & & & & 51.8 / 46.2 & \ \scriptsize{31.0 h}$^\ddagger$ & \scriptsize{43.2 G}$^\ddagger$ \\
    \end{tabular}
    \vspace{.8em}
    \caption{\textbf{Random sampling ratio for training and inference.} Numbers of cells in the upper triangular represent ``\boxAP / \maskAP''. $^\dagger$: measured with batch size of 1. $^\ddagger$: measured on A100 GPUs.}
    \label{tab: rand sample ratio}
\vspace{-1.5em}
\end{table}

\paragraph{Sampling Ratio and Type for ViT Encoder.}
We study different random sampling ratio combinations for training \& inference in Table~\ref{tab: rand sample ratio}.
In general, our \model{} can achieve \textit{better} performance if the inference sampling ratio is \textit{higher} than the training sampling ratio.

Specifically, under the scenario of inference with full input, we find with only 25\% of the input for the ViT encoder during training, \model{} can already achieve a very compelling performance that outperforms the upgraded Swin Transformer (detailed in \cref{sec: main results} and Table~\ref{tab: main result}).

Furthermore, with 50\% of the input for the ViT encoder during training, \model{} obtains very competitive results which are similar to the 100\% input training \& inference setting, striking a good trade-off between accuracy and resources.
From another perspective, in Table~\ref{tab: additional ft} we show that adding an additional short fine-tuning stage (6-epoch) with full inputs after the fine-tuning procedure (36-epoch) with sampled inputs does \textit{not} help further improve the accuracy.
These results indicate that training with 50\% randomly sampled inputs is sufficient to achieve a satisfactory performance.

The compelling results of using only 50\% input for training are not a coincidence, as we believe it is closely related to the use of ConvStem.
Specifically, the receptive field size of our ConvStem is 31, approximately 2$\times$ of the stem's output (also the ViT encoder's input) feature's stride.
That means if we sample with a stride of 2 on the input feature of ViT encoder (\ie{}, grid sampling with a ratio of 50\%), the sampled feature map can almost cover all locations of the original input image.
In practice, we find uniform random sampling generalizes much better than grid sampling in the full input set inference scenario, as shown in Table~\ref{tab: rand or grid}.


\begin{table*}[t]
\vspace{-.5em}
\centering
\begin{minipage}{0.45\linewidth}{\begin{center}
\tablestyle{4pt}{1.05}
\begin{tabular}{x{42}x{42}|x{28}x{28}}
1$^{\text{st}}$-stage ft & 2$^{\text{nd}}$-stage ft & \boxAP & \maskAP \\
\shline
\baseline{50\% rand} & \baseline{\xmark} & \baseline{51.5} & \baseline{46.0} \\
50\% rand & full set & 51.4 & 46.0 \\
\end{tabular}
\captionsetup{width=.95\linewidth}
\caption{\textbf{Study of additional fine-tuning with full input set.} Fine-tuning once with a sampling ratio of 50\% is sufficient.}
\label{tab: additional ft}
\end{center}}
\end{minipage}
\hspace{2em}
\begin{minipage}{0.45\linewidth}{\begin{center}
\tablestyle{4pt}{1.05}
\begin{tabular}{y{42}x{42}|x{28}x{28}}
\ \ \ training & inference & \boxAP & \maskAP \\
\shline
\baseline{\ 50\% rand} & \baseline{full set} & \baseline{51.5} & \baseline{46.0} \\
\ 50\% grid & full set & 48.7 & 44.0 \\
\end{tabular}
\captionsetup{width=.95\linewidth}
\caption{\textbf{Random sampling \vs grid sampling.} Random sampling generalizes well in the full input set inference scenario.}
\label{tab: rand or grid}
\end{center}}
\end{minipage}
\vspace{-1.5em}
\end{table*}

\paragraph{Inference Strategy.}

\begin{table}[ht]
\vspace{-.5em}
    \centering
    \tablestyle{4pt}{1.2}
    \begin{tabular}{@{}z{60}|x{56}x{56}x{56}x{56}x{56}}
        \# eval $\times$ ratio & \baseline{1$\times$100\%} & 1$\times$50\% rand & 2$\times$50\% rand & 4$\times$50\% rand & 8$\times$50\% rand \\
        \shline
        \boxAP / \maskAP & \baseline{51.5 / 46.0} & 49.5 / 44.3 & 50.3 / 44.8 & 50.9 / 45.3 & 51.0 / 45.2 
    \end{tabular}
    \vspace{.8em}
    \caption{\textbf{Study of different inference strategies.} ``\# eval $\times$ ratio'': the ensemble result of several independent evaluations with a specific sampling ratio, \eg, ``2$\times$50\% rand'' means the ensemble accuracy of inference twice with a random sampling ratio of 50\%.}
    \label{tab: inference strategy}
    \vspace{-1.em}
\end{table}

We study the appropriate inference strategy for \model{} in Table~\ref{tab: inference strategy}.
Under the scenario of inference with sampled inputs, we find the ensemble of more trials generally improves the performance compared with inference only once with sampling.
Meanwhile, inference with the full input set, which can be also regarded as \textit{an ensemble of input features} for ViT encoder since it is pre-trained with only partial observations, is more performant.
These results imply that the ensemble of input features works better than the ensemble of output results for our approach.

To summarize, results in Table~\ref{tab: rand sample ratio}, \ref{tab: additional ft}, \ref{tab: rand or grid} and \ref{tab: inference strategy} indicate that our \colorbox{baselinecolor}{default} training and inference strategies are nearly optimal.

\paragraph{ConvStem Type.}
In Table~\ref{tab: convstem}, we study whether increasing the capacity of the convolutional earlier stage is beneficial to the detection performance, and the answer is negative.
We choose the recent state-of-the-art ConvNeXt's~\cite{convnext} earlier stage design\footnote{We directly adopt the configuration / pre-trained weights of the first two stages of ConvNeXt-Large, since this design has matched model size as well as output channels.} as a replacement of our default / na\"ive design, and we observe no further improvement.
This implies that for the earlier stage, its convolutional property matters more than its strength.
As the FPN can help fuse different-level features to have strong semantics at all scales~\cite{FPN}, the expressiveness of high-resolution features for object-level recognition shouldn't be a worry.

\paragraph{Decoder Input Feature.}
Since we only encode partial inputs, to obtain the full set of features, we need to fill in all unsampled locations for the decoder as well as the task layer, as studied in Table~\ref{tab: mask or conv}.
One straightforward solution is to fill the blank with $\mathtt{[MASK]}$ tokens, as the decoder is pre-trained to process them.
While we find using the ConvStem output feature is slightly better.

\begin{table*}[!t]
\vspace{-1.em}
\centering
\begin{minipage}{0.45\linewidth}{\begin{center}
\tablestyle{4pt}{1.05}
\begin{tabular}{x{54}x{32}|x{28}x{28}}ConvStem & pt? & \boxAP & \maskAP \\
\shline
\baseline{na\"ive} & \baseline{\xmark} & \baseline{51.5} & \baseline{46.0} \\
ConvNeXt~\cite{convnext} & \xmark & 51.3 & 45.8 \\
ConvNeXt~\cite{convnext} & \cmark & 51.4 & 45.8 \\
\end{tabular}
\captionsetup{width=.95\linewidth}
\caption{\textbf{Study of the ConvStem type.} The na\"ive design (~\cref{sec: convstem}) with random initialization is sufficient.}
\label{tab: convstem}
\end{center}}
\end{minipage}
\hspace{2em}
\begin{minipage}{0.45\linewidth}{\begin{center}
\tablestyle{4pt}{1.05}
\begin{tabular}{x{64}|x{28}x{28}}
dec input feat & \boxAP & \maskAP \\
\shline
\baseline{$\mathtt{[MASK]}$} token & \baseline{51.5} & \baseline{46.0} \\
ConvStem feat & 51.7 & 46.1 \\
\multicolumn{3}{c}{~}\\
\end{tabular}
\captionsetup{width=.95\linewidth}
\caption{\textbf{Study of the decoder input feature type of all unsampled positions.} ConvStem feature is slightly better.}
\label{tab: mask or conv}
\end{center}}
\end{minipage}
\vspace{-.5em}
\end{table*}

Using ConvStem features at unsampled locations is an analogy of \textit{stochastic depth}~\cite{droppath}, as we ``randomly drop a subset of layers (\ie, all unsampled embeddings of the ViT encoder) and bypass them with the identity function (\ie, ConvStem features)~\cite{droppath}.''
Therefore, this training and inference strategy also aligns with the motivation of ``train \textit{short} networks and use \textit{deep} networks at test time.'' as well as ``an implicit \textit{ensemble} of network of different depths~\cite{droppath}.''
Since using the ConvStem feature brings nearly cost-free improvement, we use it for comparisons with other leading approaches in Table~\ref{tab: main result}.

\paragraph{Number of Decoder Layers.}

\begin{table}[ht]
    \tablestyle{8pt}{1.1}
    \begin{tabular}{@{}c|ll|ccc@{}}
        \# dec layers & \multicolumn{1}{c}{AP$^\text{box}$} & \multicolumn{1}{c|}{AP$^\text{mask}$} & \# params & training mem$^\dagger$ & training time \\
        \shline
        1 & 49.9             & 44.6             & 1.00$\times$ {\scriptsize (118 M)} & 1.00$\times$ {\scriptsize (12.2 G)} & 1.00$\times$ {\scriptsize (15.5 h)} \\ 
        2 & 50.6 & 45.2 & 1.03$\times$ {\scriptsize (121 M)} & 1.18$\times$ {\scriptsize (14.4 G)} & 1.12$\times$ {\scriptsize (17.3 h)} \\
        \baseline{4} & \baseline{51.5} & \baseline{46.0} & \baseline{1.08$\times$ {\scriptsize (127 M)}} & \baseline{1.63$\times$ {\scriptsize (19.9 G)}} & \baseline{1.16$\times$ {\scriptsize (20.3 h)}} \\
        8 & 51.6 & 46.1 & 1.18$\times$ {\scriptsize (140 M)} & 2.39$\times$ {\scriptsize (29.2 G)} & 1.74$\times$ {\scriptsize (27.0 h)} \\
    \end{tabular}
    \vspace{.8em}
    \caption{\textbf{Study of the number of decoder layers.} $^\dagger$: measured with batch size of 1.}
    \label{tab: number of dec}
\vspace{-2.5em}
\end{table}

We study the impact of the number of MAE pre-trained decoder layers in Table~\ref{tab: number of dec}.
The lightweight decoder is used to recover full features given encoded partial observations.
We find a decoder with 4 layers achieves the best trade-off.
If we randomly initialize the decoder, the training diverges using our default configuration.

It is usually believed that the MAE encoder learns general representations that transfer well, while by analogy with BERT~\cite{BERT} from NLP, we believe the MAE decoder is actually trained to be a BERT encoder, and the MAE encoder is more like a BERT tokenizer that maps the raw input signal to BERT encoder's input embeddings. 
To our knowledge, \model{} is the first work to leverage the MAE pre-trained decoder in downstream tasks.
What the MAE decoder learns during pre-training is still unclear, and the property of it deserves more attention in future research.

\paragraph{ViT's \& FPN's Input Form.}

\begin{table}[ht]
\vspace{-.5em}
    \centering
    \tablestyle{2pt}{1.1}
    \begin{tabular}{@{}z{24}|x{52}x{52}x{52}|x{32}x{32}|x{32}x{32}}
        & hybrid FPN? & rand sample? & \# dec layers & \boxAP & \maskAP & \# params & time\\
        \shline
        \baseline{row1} & \baseline{\cmark} & \baseline{\cmark} & \baseline{4} & \baseline{51.5} & \baseline{46.0} & \baseline{\scriptsize{127 M}} & \baseline{\scriptsize{20.3 h}} \\
        row2 & \cmark & \cmark & 1 & 49.9 & 44.6 & \scriptsize{118 M} & \scriptsize{15.5 h} \\
        row3 & \cmark & \xmark & 0 & 49.5 & 44.5 & \scriptsize{113 M} & \scriptsize{17.5 h} \\
        row4 & \xmark & \xmark & 0 & 48.9 & 43.9 & \scriptsize{120 M} & \scriptsize{18.4 h} \\
    \end{tabular}
    \vspace{.8em}
    \caption{\textbf{Study of the ViT's \& FPN's input form.} ``hybrid FPN?'': the FPN's $P_2$ \& $P_3$ features come from ConvStem's intermediate features (\cmark) or ViT's intermediate features~\cite{xcit, benchmarking} (\xmark). ``rand sample'': the ViT encoder's input comes from the 50\% randomly sampled ConvStem's output (\cmark) or window partitioned ConvStem's output with a window size of 7$\times$7~\cite{Swin, benchmarking} (\xmark).}
    \label{tab: vit fpn input}
    \vspace{-1.5em}
\end{table}

In Table~\ref{tab: vit fpn input}, we conduct ablation studies on \model{}-Base with only one decoder layer (row2) to align with the budgets of row3 and row4.

row2 \& row3 show that randomly sampled inputs can achieve better performance than window partitioned inputs~\cite{Swin, benchmarking} in \model{} given similar budgets.
Also notice that row3 is a counterpart of Swin Transformer\footnote{The \#Transformer layers of different stages in Swin-Base is \{2, 2, 18, 2\}. Therefore both Swin-Base and \model{}-Base in row3 of Table~\ref{tab: vit fpn input} have a deep \& strong 3$^{rd}$-stage with window attention, as well as a shallow \& compact early stage.}.
Compared with the well-established Swin+ in Table~\ref{tab: main result} (49.2 \boxAP / 43.5 \maskAP), row3 with a ConvNet (rand.init.) \& ViT (pre-trained) hybrid architecture can obtain higher accuracy, indicating that pre-training specific feature extractor for visual recognition may no longer be indispensable given the strong representation inside vanilla ViT.

row3 $\to$ row4 demonstrates that the performance will suffer if the higher resolution inputs of FPN are from the intermediate features of ViT encoder~\cite{xcit, benchmarking} instead of ConvStem, which implies the convolutional feature is more beneficial to object-level understanding tasks. 
We show some qualitative results in the next section that can help us gain an intuitive sense.

\subsection{Visualization}
\label{sec: feat vis}

\begin{figure}[t!]
  \centering
  \includegraphics[width=1\columnwidth]{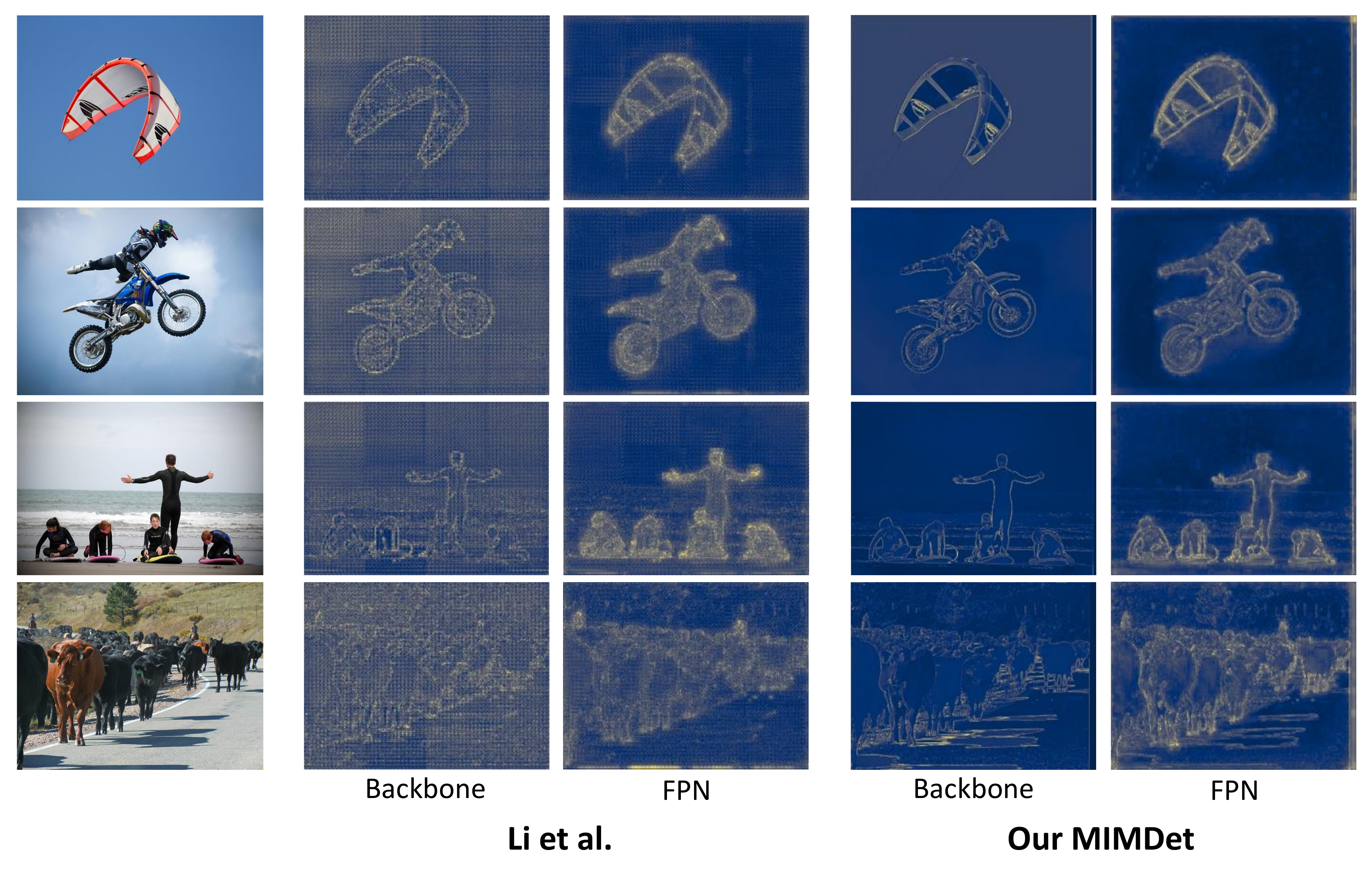}
  \caption{\textbf{Visualizations and comparisons of some stride-4 backbone and FPN feature maps.} The feature maps of \citet{benchmarking} is obtained from our re-implementation which successfully reproduces its reported results.}
  \vspace*{-0.2cm}
  \label{fig: feat vis}
\end{figure}

Figure~\ref{fig: feat vis} visualizes some backbone \& FPN feature maps with a stride of 4 for both \citet{benchmarking} and our \model{}.
The stride-4 backbone feature of \citet{benchmarking} is obtained from a stride-16 ViT encoder feature via upsampling using two stride-2 transposed convolutions with  2$\times$2 kernel.
The resulting features suffer from very strong ``checkerboard artifacts~\cite{artifacts}''.
If we look closer, the evidence of ViT attention's window partition emerges.
Thanks to FPN, the noise can be mitigated to some extent.
However, many low-level details are still fuzzy.
On the other hand, our ConvStem in \model{} can always produce clear and tidy features, which is beneficial to both the ViT encoder as well as the Mask R-CNN detector.
More visualizations are available in the Appendix.

\subsection{Comparisons with Previous Approaches}
\label{sec: main results}


Table~\ref{tab: main result} shows the comparisons.
Our \model{} can successfully adapt a MAE pre-trained representation to achieve strong \boxAP and \maskAP, better than some representative well-established hierarchical ViT architectures~\cite{Swin, MViTv2} under the same fine-tuning procedure.
\model{}-Base (Table~\ref{tab: vit-b}) without relative position biases can outperform the previous best adapted vanilla ViT~\cite{benchmarking} given the same initial representation~\cite{mae} with a more modest data augmentation strategy (resize \& crop \textit{v.s.} LSJ), while only requiring 2.8$\times$ less epochs (36 ep \textit{v.s.} 100 ep) to coverage and being 2.7$\times$ faster (20.3 h \textit{v.s.} 54.0 h) in training\footnote{Measurements and comparisons of the training time is based on our re-implementation of \citet{benchmarking}, which successfully reproduces its reported results (ViT-Base: Our reproduced results: 50.4 \boxAP / 44.9 \maskAP \textit{v.s.} \citet{benchmarking}'s: 50.3 \boxAP / 44.9 \maskAP).}.
Moreover, we observe a promising scaling trend of our approach, \ie{}, our \model{}-Large (Table~\ref{tab: vit-l}) without relative position biases can achieve 54.3 \boxAP / 48.2 \maskAP, outperforming the best large model in \citet{benchmarking} as well as strong hierarchical competitors~\cite{MViTv2}.

\begin{table*}[t]
\vspace{-.2em}
\centering

\subfloat[
Results of \textit{base-sized} models.
\label{tab: vit-b}
]{
\centering
\begin{minipage}{0.99\linewidth}{\begin{center}
    \tablestyle{2pt}{1.1}
    \begin{tabular}{@{}y{68}|x{48}x{48}x{56}x{48}|x{36}x{36}}
        backbone & pt config & ft epochs & data aug & rel pos? & \boxAP & \maskAP \\
        \shline
        \multicolumn{3}{@{}l}{\hvits \emph{\scriptsize representative hierarchical architecture}} \\
        \hline
        Swin+~\cite{Swin} & sup{\scriptsize-1K} & 36 & resize \& crop & \cmark & 49.2 & 43.5 \\
        MViTv2~\cite{MViTv2} & sup{\scriptsize-1K} & 36 & resize \& crop & \cmark & 51.0 & 45.7 \\
        \hline
        \multicolumn{3}{@{}l}{\vvits \emph{\scriptsize adapted vanilla Vision Transformer}} \\
        \hline
        \citet{benchmarking} & {\scriptsize MAE-1K} & 100 & LSJ$_\text{1024}$ & \cmark & 50.3 & 44.9 \\
        \textbf{\model{} (Ours)} & {\scriptsize MAE-1K} & 36 & resize \& crop & \xmark & \textbf{51.7} & \textbf{46.1} \\
    \end{tabular}
\end{center}}\end{minipage}
}
\\
\centering
\vspace{-.5em}
\subfloat[
Results of \textit{large-sized} models.
\label{tab: vit-l}
]{
\begin{minipage}{0.99\linewidth}{\begin{center}
    \tablestyle{2pt}{1.1}
    \begin{tabular}{@{}y{68}|x{48}x{48}x{56}x{48}|x{36}x{36}}
        backbone & pt config & ft epochs & data aug & rel pos? & \boxAP & \maskAP \\
        \shline
        \multicolumn{3}{@{}l}{\hvits \emph{\scriptsize representative hierarchical architecture}} \\
        \hline
        MViTv2~\cite{MViTv2} & sup{\scriptsize-1K} & 36 & resize \& crop & \cmark & 51.8 & 46.2 \\
        MViTv2~\cite{MViTv2} & sup{\scriptsize-21K} & 36 & resize \& crop & \cmark & 52.7 & 46.8 \\
        \hline
        \multicolumn{3}{@{}l}{\vvits \emph{\scriptsize adapted vanilla Vision Transformer}} \\
        \hline
        \citet{benchmarking} & {\scriptsize MAE-1K} & 100 & LSJ$_\text{1024}$ & \cmark & 53.3 & 47.2 \\
        \textbf{\model{} (Ours)} & {\scriptsize MAE-1K} & 36 & resize \& crop & \xmark & \textbf{54.3} & \textbf{48.2} \\
    \end{tabular}
\end{center}}\end{minipage}
}
\vspace{-.1em}
\caption{\textbf{COCO object detection and instance segmentation results.} We use Mask R-CNN as the task layer. ``rel pos'': using relative position biases~\cite{rel_pos, relpos}, which usually improves $\sim$1 AP but severely adds training time and memory. ``Swin+'': its Mask R-CNN is augmented following~\citet{benchmarking}\protect\footnotemark. ``LSJ$_\text{1024}$'': large scale jittering~\cite{lsj} on a 1024$\times$1024 canvas. ``sup{\scriptsize-21K}'': pre-training using ImageNet-21K~\cite{ImageNet-21k} with supervision.}
\label{tab: main result}
\vspace{-1.5em}
\end{table*}

\footnotetext{In previous literature~\cite{MViTv2, mpvit}, Swin-Base with standard Mask R-CNN obtains 48.5 \boxAP / 43.4 \maskAP.}

\subsection{Limitation and Discussion}
\label{sec: limitation and discussion}
A potential limitation is we only study one representative framework of MIM, \ie{}, the MAE family~\cite{mae} based on an asymmetric encoder-decoder design that allows the encoder to process only partial observations without $\mathtt{[MASK]}$ tokens during pre-training.
Nevertheless, we believe MAE is a simple, strong and scalable MIM framework that would dominate visual pre-training over the next few years as BERT~\cite{BERT} in NLP.
The MAE framework is also compatible with other advances in MIM pre-training~\cite{maskfeat, data2vec, peco}, and there is a lot of evidence that MAE generalizes well to 2D visual multi-modal \& multi-task~\cite{multimae}, medical~\cite{medmae}, video~\cite{videomae}, 3D~\cite{pointmae}, RL~\cite{rlmae} and even language~\cite{should15} \& audio~\cite{audiomae} pre-training.
Therefore we believe this work can contribute to a board research related to MIM / MAE study.



\paragraph{Reproducibility.}
The code needed to reproduce the experimental results is in the supplemental material.
We observe $\sim$0.2 AP fluctuation with respect to different random seeds, which is very common for the COCO dataset.
We report the key results using the median of 3 independent runs.

\section{Related Work}

\paragraph{Hierarchical Backbone for Object Detection.}
Well-established object detectors~\cite{FPN, RetinaNet, fcos, EfficientDet, Faster-R-CNN, MaskRCNN} usually take advantage of multi-scale features as inputs for better performance.
The multi-scale inputs naturally exist in hierarchical ConvNet~\cite{AlexNet, VGG, ResNet, resnext, EfficientNet} / ViT~\cite{Swin, PVT, MViT, MViTv2, CoaT} backbones.
This work aims to adapt and unleash MIM pre-trained vanilla ViT's representations~\cite{beit, mae, simmim, ibot} for object-level recognition without modifying its pre-training process and its architectural nature.

\paragraph{Taming Vanilla ViT for Object Detection.}
Since the introduction of Transformer~\cite{Transformer} to computer vision~\cite{ViT}, the effort of taming pre-trained vanilla ViT for object detection has never stopped.
\citet{ViT-RCNN} is the first to adapt a supervised pre-trained ViT for object detection with a Faster R-CNN detector~\cite{Faster-R-CNN}.
YOLOS~\cite{YOLOS} proposes to perform object detection in a pure sequence-to-sequence manner with a pre-trained ViT encoder only. 
Similar to YOLOS, we also treat the inputs for ViT encoder as 1D sequences instead of 2D grids.
\citet{benchmarking} is the first work to conduct a large-scale study of vanilla ViT on object detection with powerful MIM pre-trained representations~\cite{beit, mae}, demonstrating the promising capability and capacity of vanilla ViT in object-level recognition.
UViT~\cite{uvit} is a recent single-scale ViT detector with a \textit{detection-oriented} design.
Different from UViT, we aim to leverage the \textit{general} representations from MIM for high-performance object detection.
Another series of work~\cite{Up-detr, detreg} explore the pre-training of DETR framework~\cite{DETR, Deformable-DETR}.


\section{Conclusion}

In this paper, we explore how to unlock the potential of MIM pre-trained vanilla ViT for high-performance object detection and instance segmentation.
The satisfactory results imply designing and pre-training specific feature extractors for visual recognition may no longer be a \textit{sine qua non} process for computer vision research.
As vanilla ViT demonstrates extremely strong model capacity, future state-of-the-art visual recognition systems shall learn to tame it and unleash it.
These trends have already been witnessed in NLP~\cite{GPT, BERT, t5, CLIP}, and we hope our work can encourage the vision community to explore the powerful general visual representations hidden in the vanilla ViT.

{
\renewcommand{\ttdefault}{cmtt}
\bibliographystyle{plainnat}
\bibliography{neurips_2022}
}

\appendix

\section{Appendix}



\subsection{Implementation Details}
\label{app: implementation details}
\renewcommand{\ttdefault}{cmtt}  %
\begin{algorithm}[!ht]
\caption{\small ConvStem for ViT-Base (PyTorch Style), which can help preserve low-level details, produce higher resolution hierarchical features for FPN, and introduce 2D inductive biases for the ViT encoder \& detector.}
\label{arch: convstem}
\definecolor{mygray}{gray}{0.35}
\definecolor{codeblue}{rgb}{0.25,0.5,0.5}
\definecolor{codekw}{rgb}{0.85, 0.18, 0.50}
\lstset{
  backgroundcolor=\color{white},
  basicstyle=\fontsize{8.0pt}{8.0pt}\color{mygray}\ttfamily\selectfont,
  columns=fullflexible,
  breaklines=true,
  captionpos=b,
  commentstyle=\fontsize{8.0pt}{8.0pt}\color{codekw},
  keywordstyle=\fontsize{8.0pt}{8.0pt}\color{codekw},
}
\begin{lstlisting}[language=python]
# Number of Parameters: 4.1M.
ConvStem(           
  ModuleList(
    (0): Sequential(
      (0): Conv2d(3, 96, kernel_size=(3, 3), stride=(2, 2), padding=(1, 1), bias=False)
      (1): LayerNorm2d(96, eps=1e-06, affine=True) & GELU()
    )
    (1): Sequential(
      (0): Conv2d(96, 192, kernel_size=(3, 3), stride=(2, 2), padding=(1, 1), bias=False)
      (1): LayerNorm2d(192, eps=1e-06, affine=True) & GELU() # Input for FPN P2.
    )
    (2): Sequential(
      (0): Conv2d(192, 384, kernel_size=(3, 3), stride=(2, 2), padding=(1, 1), bias=False)
      (1): LayerNorm2d(384, eps=1e-06, affine=True) & GELU() # Input for FPN P3.
    )
    (3): Sequential(
      (0): Conv2d(384, 768, kernel_size=(3, 3), stride=(2, 2), padding=(1, 1), bias=False)
      (1): LayerNorm2d(768, eps=1e-06, affine=True) & GELU()
      (2): Conv2d(768, 768, kernel_size=(1, 1), stride=(1, 1)) # Input for ViT-Base Enc.
    )))
\end{lstlisting}
\end{algorithm}
\paragraph{Architecture of ConvStem.}
We adopt a minimalist ConvStem design, \ie{}, by simply stacking 3$\times$3 regular convolutions with a stride of 2 and doubled feature dimensions.
Each convolutional layer is followed by a layer normalization~\cite{LN} and a GELU activation~\cite{GELU}.
The detailed configurations are given in Architecture~\ref{arch: convstem}.

\paragraph{Hyper-parameters and Model Configurations.}
Hyper-parameters and model configurations for fine-tuning on the COCO dataset are shown in Table~\ref{app: hyper & config}.
Since the vanilla ViT encoder is already pre-trained while the task layer is trained from scratch, the learning rate of the ViT encoder part is divided by a ``lr multiplier'' and the learning rate for the task layer is multiplied by a ``lr multiplier''.

\renewcommand{\ttdefault}{ptm}
\newcommand{\lr}{\emph{lr}\xspace}
\newcommand{\wtd}{\emph{wd}\xspace}
\newcommand{\drp}{\emph{dp}\xspace}
\newcommand{\expnum}[2]{{#1}\mathrm{e}^{#2}}
\begin{table}[h]
    \tablestyle{6pt}{1.05}
    \vspace{-.5em}
    \begin{tabular}{@{}l|ccccc|ccc@{}}
    & \multicolumn{5}{c|}{hyper-parameters} & \multicolumn{3}{c}{model configs} \\
    	backbone & \scriptsize{lr} & \scriptsize{lr multiplier} & \scriptsize{weight decay} & \scriptsize{drop path} & \scriptsize{ft epochs} & \scriptsize{params (M)} & \scriptsize{FLOPs (G)} & \scriptsize{inf. time (s)} \\
    	\shline
    	\model{}-Base & $\expnum{8}{-5}$ & 2 & 0.1 & 0.1 & 36 & 128 & 933 & 0.29 \\
    	\model{}-Large & $\expnum{8}{-5}$ & 3.5 & 0.1 & 0.1 & 36 & 349 & 2082 & 0.58 \\
    \end{tabular}
    \vspace{.5em}
    \caption{Hyper-parameters and and model configurations for COCO fine-tuning. We report the average number of FLOPs and inference time for the first 100 images in the COCO $\mathtt{val}$ set following~\cite{DETR} on a V100 GPU.}
     \label{app: hyper & config}
    \vspace{-1.5em}
\end{table}
\paragraph{Optimization.}
The loss function of \model{} keeps the \textit{same} as the canonical Mask R-CNN~\cite{MaskRCNN, benchmarking}, \ie{}, explicit reconstruction loss for ViT encoder is \textit{not} needed during the fine-tuning, even though the encoder only receive partial observations.
The implicit reconstruction process of ViT encoder is driven by the supervision from the Mask R-CNN detector.

\subsection{More Visualizations}
\label{app: more visualizations}

\begin{figure}[h]
  \centering
  \includegraphics[width=1\columnwidth]{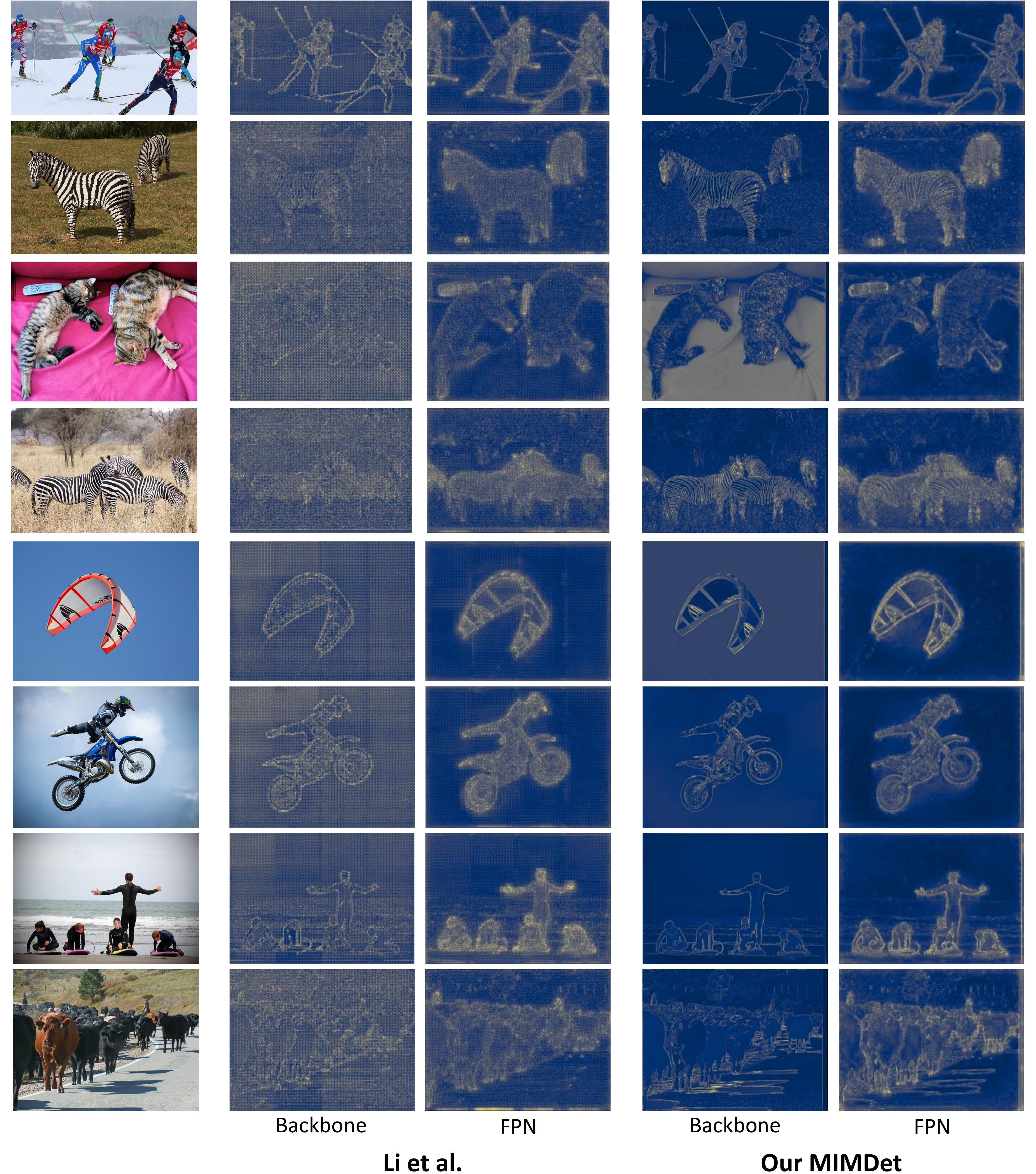}
  \caption{\textbf{More visualizations and comparisons of some stride-4 backbone and FPN feature maps.} The feature maps of \citet{benchmarking} is obtained from our re-implementation which successfully reproduces its reported results.}
  \vspace*{-0.2cm}
  \label{fig: more feat vis}
\end{figure}


\end{document}